\documentclass[conference]{IEEEtran}

\usepackage{cite}
\usepackage{amsmath,amssymb,amsfonts}
\usepackage{algorithmic}
\usepackage{graphicx}
\usepackage{textcomp}
\usepackage{xcolor}
\usepackage[]{hyperref}
\usepackage{subfigure}
\usepackage{booktabs}
\usepackage{multirow}

\graphicspath{{./figures/}}

\begin{document}

\title{Using Convolutional Neural Networks for fault analysis and alleviation in accelerator systems}

\author{\IEEEauthorblockN{Jashanpreet Singh Sraw}
\IEEEauthorblockA{\textit{Thapar Institute of Engineering and Technology}\\
Patiala, India}
\and
\IEEEauthorblockN{Deepak M C}
\IEEEauthorblockA{\textit{PES College of Engineering}\\
Mandya, India}
}

\maketitle

\begin{abstract}
    Today, Neural Networks are the basis of breakthroughs in virtually every technical domain. Their application to accelerators has recently resulted in better performance and efficiency in these systems. At the same time, the increasing hardware failures due to the latest (shrinked) semiconductor technology needs to be addressed. Since accelerator systems are often used to back time-critical applications such as self-driving cars or medical diagnosis applications, these hardware failures must be eliminated. Our research evaluates these failures from a systemic point of view. Based on our results, we find critical results for the system reliability enhancement and we further put forth an efficient method to avoid these failures with minimal hardware overhead.
    
\end{abstract}


\IEEEpeerreviewmaketitle

\section{Introduction} \label{sec:introduction}
Deep neural networks (DNNs) have been demonstrated to be successful in 
massive territories like image processing, video processing, and 
natural language processing over the years \cite{Gatys_2016_CVPR, Collobert:2008:UAN:1390156.1390177, Chen_2015_ICCV}. The success further provokes the 
flourish of customized neural network accelerators 
with massive parallel processing engines which typically offer
much higher performance and energy efficiency compared to 
general purposed processors (GPPs) 
\cite{chen2014diannao, chen2014dadiannao,  chen2016eyeriss, Zhang:2015:OFA:2684746.2689060, 7551397}.
While the performance of neural network accelerators has been intensively 
optimized from various angles like pruning and quantization, 
the reliability of the accelerators especially 
the ones that are deployed on FPGAs remains not well-explored.

The shrinking semiconductor technology greatly improves the 
transistor density of chips, but the circuits with smaller 
feature size are more susceptible to manufacturing defects and abnormal 
physical processes like thermal stress, electromigration, 
hot carrier injection, and gate oxide wear-out. Thereby, the 
persistent (hard) faults become one of the major sources 
of system unreliability. Despite the fault-tolerance of 
neural network models, permanent faults in neural network 
accelerators can cause computing errors and considerable 
wrong predictions \cite{238315}. They may even lead to disastrous 
consequences to some of the mission-critical applications such as 
self-driving, nuclear power plants, and medical diagnosis, 
which are sensitive to neural network prediction mistakes. 

It is seminal to comprehend the impact of errors on neural network accelerators to improve the model functioning. To this end, several research works are published that contain methods to identify errors in neural network accelerators \cite{Reagen2018, Kausar2016Artificial, Li2018TensorFI, Li2017, Salami2018}. However, they usually experimented with 
simulators and focused on the influence of hardware faults on the 
prediction accuracy of the neural network models. 
For instance, the authors in \cite{Reagen2018} 
mainly investigated the data faults of weights, activities, 
and hidden states, which are stored in on-chip buffers.
Then they explored the neural network model resilience with 
model-wise analysis and layer-wise analysis. The work 
in \cite{Li2017} targeted at the analysis of software errors 
in DNN accelerators and explored the error propagation 
behaviors based on the structure of the neural networks, 
data types and so on. 

The simulation based approaches are fast and flexible for analyzing 
neural network computing and prediction accuracy, but they typically 
ignore critical controlling details and interfaces of 
the DNN accelerators to ensure the simulation speed. Nevertheless,
hardware faults on these components may have considerable influence 
to the overall acceleration system other than the prediction 
accuracy loss. For instance, faults in DMA 
module may result in illegal memory accesses and corrupt the system.
This must be addressed to guarantee reliable DNN 
acceleration especially in mission-critical applications.
In addition, many DNN accelerators are implemented on FPGAs for 
more intensive customization and convenient reconfiguration. 
While the functionality of the DNN accelerators is mapped to the 
FPGA infrastructures instead of the primitive logic gates, 
the simulation based approaches that usually inject errors to the 
operations used in DNN processing do not apply to the 
FPGA based DNN acceleration system. Because hardware faults 
in FPGAs affect the configuration of the devices instead of the 
accelerator components directly while the actual parts of the 
accelerators that are influenced depends on the FPGA placing 
and routing. 

To gain further insight of faults in DNN accelerators, 
we conduct the fault analysis on running ARM-FPGA system 
where FPGA has a representative DNN accelerator with 2D systolic array 
implemented along with hardware fault injection modules and shares 
the DRAM with the ARM processors. Since the fault distribution models are implemented 
with software on the ARM processors, they are convenient to change for fault 
analysis using different models. Aside from that we work with four different models for detecting the others. 

Li, Shunlong et. al. \cite{li2021novelty} operated one-dimensional convolutional neural network along the time axis to capture the temporal dependency. Shi, Chong et. al \cite{shi2021fault}'s work focuses on the internal leakage fault diagnosis caused by the wear based on intrinsic mode functions (IMFs).Unlike prior works that focused on prediction accuracy loss analysis, 
we try to analyze the behaviours of the DNN accelerators under hardware faults and investigate the system functionality, 
fault coverage, input variation. 
Particularly, we study the system stall that dramatically destroys the system 
functionality in detail and present a simple yet efficient approach 
to alleviate the problem.

The following lines sum up our contributions to the litereature:
\begin{itemize}
	\item Our system focuses on fault analysis of neural network accelerators on ARM-FPGAs. To make the analysis easier, it provides high-level interfaces for both fault injection and output data comparison of neural network models.
	
	\item On top of the fault analysis system, we mainly classify 
	and analyze the resulting behaviors of the DNN accelerators 
	from system point of view. On top of the prediction accuracy, we 
	also study the fault coverage, system functionality, input 
	variation when the accelerators are exposed to persistent faults.
	
	\item With comprehensive experiments over representative models, we observe that system stalls that can 
	destroy the system functionality of DNN accelerators cannot be 
	ignored. And we further show that errors in data movement instructions 
	is a key reason for the system stalls, which can be addressed with 
	negligible hardware overhead.
\end{itemize}

The remainder of the paper can be described in the following paragraphs. Section \ref{sec:background} sheds light on, the introduction to deep learning neural networks and a typical DNN accelerators with 2D computing array. In Section \ref{sec:fault-analysis}, 
we describe the proposed framework on ARM-FPGAs and elaborate on the major aspects that we will investigate from system point of view. 
Section \ref{sec:experiment} details the results and evaluation. In Section \ref{sec:relatedwork}, we give a brief 
review of prior fault tolerant analysis and design of neural network accelerators.
Finally, we conclude this work in Section \ref{sec:conclusion}.

\section{Background}\label{sec:background}
The main causes of unreliability are hardware flaws in DNN accelerators. The impact of errors is inextricably linked to the microarchitecture of the DNN accelerators. In this part, we take a standard DNN accelerator with a normal 2D processing element (PE) array as an example and develop its design to assist understand and study the fault tolerance of the accelerators.

\begin{figure}
    \center{\includegraphics[width=0.8\linewidth]{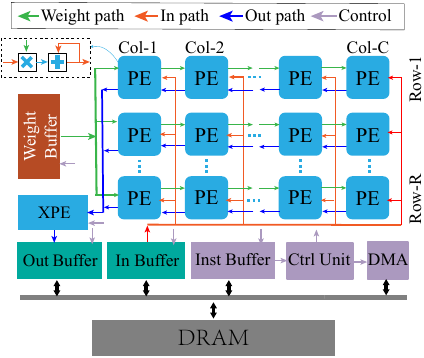}}
    \caption{Typical DNN accelerator architecture}
\vspace{-0.5em}
\label{fig:npu-arch}
\end{figure}

Figure \ref{fig:npu-arch} depicts a sample DNN accelerator. It uses output stationary data flow to transfer computations like convolution to a 2D computing array.
Every PE completes all of the processes necessary to produce an output activation. While each PE just has a single multiplier and accumulator, it progressively accumulates all input activations in a filter window. While input activations are organized in a batch and sent 
to one column of PE every cycle, each PE in the column shares the 
same input data through broadcasting and it takes 
each PE multiple cycles to complete the accumulation. 
During this period, more batched input activations 
can be read and sent to the next column of PEs along with 
the movement of the weights. Output activations flow 
from right to left in column-wise. Eventually, each row of the PEs
array produces a set of sequential output activations in 
the same row of one output feature map on y-axis. 
The architecture along with the 
compact data flow achieves high data reuse under limited 
on-chip buffer bandwidth provision. 

Both convolution layer and full connection layer can be mapped to the 
array efficiently. While pooling and other non-linear activation functions 
such as sigmoid will be performed right after the computing-intensive 
layers like convolution layer in a module named XPE such that 
the data movement between the two layers can be reduced. 
All the neural network operations can be mapped to the 
accelerator. To associate the models with various layer combinations and characteristics such as stride size and kernel size, we define 
a set of instructions to generate appropriate control signals 
for different neural network operations. Each neural network 
will be compiled to a series of instructions and executed sequentially. 
In addition, neural network input features and weights 
are usually larger than the on-chip buffers and PE array size, 
so they must be tiled and the tiles need to be scheduled to 
obtain efficient execution on the accelerator. To enable fine-grained 
optimizations, each instruction only handles operations of 
a single tile. Thus, tiling is performed during model
compilation and it is transparent to the instructions.

Table ~\ref{tab:instrction-set} shows the instruction set of the neural 
network accelerator. It adopts 64-bit fixed length encoding 
and consists of four types of instructions including parameter 
setup, calculation, data movement and control. The parameter setup 
category defines the input/output feature size, kernel size, 
Q-code, and DMA parameter. Calculation 
category includes different operations in neural networks such as 
convolution, full connection, pooling, addition, softmax, 
dot-accumulation and activation function etc. Data movement category
includes three instructions which move a block of data 
from DRAM to buffer, buffer to DRAM and buffer to buffer 
respectively. Finally, control category includes three instructions which are 
Jump, Stop and Nop. Jump instruction is mainly used for repeated 
execution. Stop is used to terminate the execution of the accelerator.
Nop is used to resolve the data dependency between sequential 
instructions.

\begin{table}
    \centering
    \caption{Instruction Set of the DNN Accelerator}
    \label{tab:instrction-set}
    \begin{tabular}{cp{0.6\columnwidth}}
        \toprule
        Instruction Type & Description \\
        \midrule
        Parameter setup & Setup parameters for the computing operations such as the input/output feature size, kernel size, Q-code, and DMA options\\
        \midrule
        Calculation & Performs various neural network operations such as convolution, full connection, pooling, addition, softmax, dot-accumulation and activation function \\
        \midrule
        Data movement & Move a block of data from buffer to buffer, buffer to DRAM and DRAM to buffer.\\
        \midrule
        Control & Control the execution of the accelerator such as Jump, Nop and Stop\\
    \bottomrule
    \end{tabular}
    \vspace{-1em}
\end{table}

The neural network accelerator architecture is general enough to support 
various neural network models. In addition, it typically works along 
with a general purposed processor and has an AXI slave port that allows 
configuration and controlling from the attached processor. It assumes 
the input data, weight and output data are stored in DRAM that can be 
accessed directly.
\section{DNN Acceleration Fault Analysis Platform and Fault Classification}
\label{sec:fault-analysis}
\subsection{Fault Analysis Platform}
\begin{figure}
    \center{\includegraphics[width=0.99\linewidth]{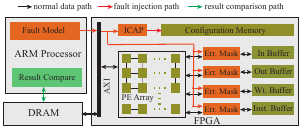}}
    \caption{Overview of the fault analysis system}
\vspace{-0.5em}
\label{fig:fault-analysis-overview}
\end{figure}

We built a fault analysis platform as represented in Figure \ref{fig:fault-analysis-overview}. It has an unusual neural network accelerator based on FPGA.

In addition to this system, we needed a module as present in Figure 2. 
The fault injection data path 
is marked with orange arrows. It is implemented on both the ARM processor 
and FPGA. The FPGA errors are prevalent in the four different memory types. 

For FPGA configuration memory, we leverage Xilinx ICAP 
port \cite{UG953}, that allows user logic to access configuration 
memory, to inject errors. We select the frames and the number of bits for each 
bit error injection arbitrarily. It is possible that the error may be present at any location of the FPGA configuration memory.
After finding the error location, we read the whole frame 
out of the configuration memory \cite{PG134}, 
change the victim bit in the frame, and write it back to the configuration 
memory \cite{UG470}. 

\begin{figure}
    \center{\includegraphics[width=0.7\linewidth]{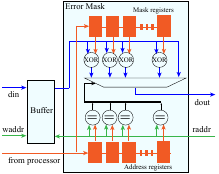}}
    \caption{Error mask for fault injection to on-chip memory}
\vspace{-0.5em}
\label{fig:error-mask}
\end{figure}

We also produce an injection mask for block RAM and distributed memory as represented in Figure \ref{fig:error-mask}. It essentially consists of a collection of address and mask registers. The addresses in the registers reflect the positions of the faults to be injected, while the associated masks keep track of the precise error bits. 

Another important part of the fault analysis platform is the result 
collection and comparison. We have the output of neural networks 
stored in the shared DRAM, which can be easily accessed on the ARM processor 
and compared to the pre-calculated golden reference. In addition, the computing 
results produced in the intermediate layers can also be compared.
The result comparison is mainly performed on the ARM processor and can be 
easily utilized for further analysis by high-level application designers.

\subsection{System fault classification}
Prior fault analysis works usually focus on computing errors of the neural networks 
and the incurred prediction accuracy loss. We argue that the consequences 
of hardware faults on a DNN acceleration system vary and should be classified 
into more categories. Table \ref{tab:classification} shows the proposed classification.
From the perspective of a system, the consequences caused by hardware faults 
roughly include system exception and accuracy degradation. System exception 
indicates that the neural network execution behaves abnormally, 
which may be stalled without returning or returns too fast or too slow.
Basically, we define them as system stall and abnormal runtime.
For the accuracy degradation, we
further classified into two cases including system stall 
and abnormal runtime.

\begin{table}
    \centering
    \caption{Hardware Faults}
    \label{tab:classification}
    \begin{tabular}{c|cccc}
    \hline
      \multirow{2}{*}{system exception} & system stall\\
      \cline{2-2}
      & abnormal runtime\\
      \cline{1-2}
      \multirow{3}{*}{Accuracy degradation} &$L_{0}$\\
      \cline{2-2}
      & ... \\
      \cline{2-2}
      &$L_{k}$\\
      \cline{1-2}
      \hline
    \end{tabular}
\end{table}

We chose Neural Networks in four different application 
scenarios and try to analyze the differences of error 
tolerance in different application scenarios and networks. 
The four network applications include ResNet network for 
image classification, YOLO system for target detection, 
LSTM network for voice classification, and DCGAN network 
for image generation. We will evaluate their fault 
tolerance from accuracy and output consistency.

Neural Network accelerators injected with hardware errors 
may produce unexpected conditions. We define the system 
halt situation, which refers to a serious error in the 
system, or working improperly. such as unable to read and 
write registers, timeout, abnormal short runtime, etc. When 
the system halts, we need to reset the FPGA and restart the 
system. System halt situations are considered the result of 
errors in the evaluation of network accuracy and are listed 
separately in the output consistency.

Network accuracy refers to the overall accuracy of the 
network when performing corresponding tasks, such as the 
accuracy of 20,000 image recognition. When there are errors 
in the operation, the accuracy of the network will show a 
downward trend. For classification networks including 
ResNet and LSTM, top-5 accuracy is used to evaluate their 
accuracy, and for the YOLO system, mAP is adopted.

Output consistency is the difference between the result of 
running with errors injected and the result of normal 
running. We ran the corresponding data set when no errors 
are injected into each network at first, and defined the 
results as standard output. The results of Neural Networks 
prediction injected with errors are divided into two 
categories: result with deviation and result match. Result 
with deviation refers to system works properly with output 
differently from standard output. Result match means that 
the system works properly and the outputs are still 
standard output.

For the result with deviation case, we define its deviation 
quantitatively and further subdivide the result. Due to the 
different application functions of each network, the 
evaluation criteria of deviation are also different. For 
YOLO system, the result is the target detection bounding 
box, and when the detection result does not match the 
standard output in object type, it is defined as detection 
result error. When the result target type is consistent, 
the error is defined as the intersection area of the error 
output and the standard output divided by the area of the 
union. Target types not match for one single level, the two 
do not overlap with each other for one level, and then each 
20\% is divided into one level. For ResNet and LSTM, the 
outputs are top-5 labels. When the error output is not 
completely consistent with the standard output, the number 
of elements in the intersection of the two is taken, and 
divide the levels refer to the number. For DCGAN, we used 
the universal SSIM standard, and divided it into six levels 
according to the actual visual effects: 0~10\%, almost 
impossible to recognize; 10~20\%, barely visible; 20~50\%, 
with large deformation or distortion; 50~80\% with partial 
deformation or distortion; 80~90\%, small deformation or 
distortion can be seen; 90~100\%, almost no deformation or 
distortion is visible.

\section{Experiment} \label{sec:experiment}
We seek to understand the influence of persistent errors on 
FPGA-based neural network acceleration system. Particularly, 
we try to analyze the influence from a system point of view
and figure out the underlying reasons for severe system problems 
such as system stall and dramatic prediction accuracy loss. 

\subsection{Device and Environment}
Xilinx Zynq-7000 SoC ZC706 Evaluation Board will be used in 
the experiments’ hardware implementation. It has 
appropriate hardware resources and is easy to develop and 
use. The hardware design and Bitstream file compilation 
were completed using the upper computer with Intel Core 
i7-6700 processor and 2x8GB DDR4 2400MHz memory. The system 
environment used was Ubuntu 16.04 LTS version, and Xilinx 
Vivado Design Suite and Xilinx Software Development Kits 
version 2017.4.

The hardware resource utilization of the error analysis 
platform implementation on ZC706 is shown in Table \ref{tab:utilization report}.

The four models cover a broad range of applications.
Yolo represent a typical neural network model for object detection \cite{redmon2016yolo9000}, 
Resent is a widely adopted neural network model for classification \cite{He_2016_CVPR}, 
LSTM is the mostly used neural network model for audio classification 
tasks \cite{sak2014long}, and DCGAN stands for a typical neural network model for 
generative tasks \cite{radford2015unsupervised}. 

Despite the widespread adoption of deep learning 
neural network on various applications, it is particularly 
successful in four categories of tasks including object detection, 
object classification, voice recognition and style transfer.
Among a great number of neural network models,
Yolo, Resnet, LSTM and DCGAN are four typical neural networks 
that are comprehensively explored to handle the four computing 
tasks respectively. 

\begin{table}
    \centering
    \caption{Utilization Report of Fault Analysis Platform}
    \label{tab:utilization report}
    \begin{tabular}{cccc}
        \toprule
        Resource & Utilization & Available & Utilization Percentage \\
        \midrule
            LUT & 122618 & 218600 & 56.09\% \\
            LUTRAM & 185 & 70400 & 0.26\% \\
            FF & 84641 & 437200 & 19.36\% \\
            BRAM & 203 & 545 & 37.25\% \\
            DSP & 297 & 900 & 33\% \\
            MMCM & 1 & 8 & 12.5\% \\
        \bottomrule
    \end{tabular}
\vspace{-1em}
\end{table}

\subsection{Overview}
We hope to explore the possible consequences of errors, the 
fault-tolerant ability of different applications and the 
influence of different error locations of accelerators, 
which can provide meaningful references for the subsequent 
network optimization and the fault-tolerant design of 
accelerators.

We conducted five sets of experiments, which explained the 
influence of single error on different networks, the 
difference of fault tolerance ability of different 
applications, the error classification of different 
applications, the influence of different accelerator units 
and whether the input data affected the error representation.

\subsection{Error Consequences and Coverage}
\begin{figure}
	\center{\includegraphics[width=0.9\linewidth]{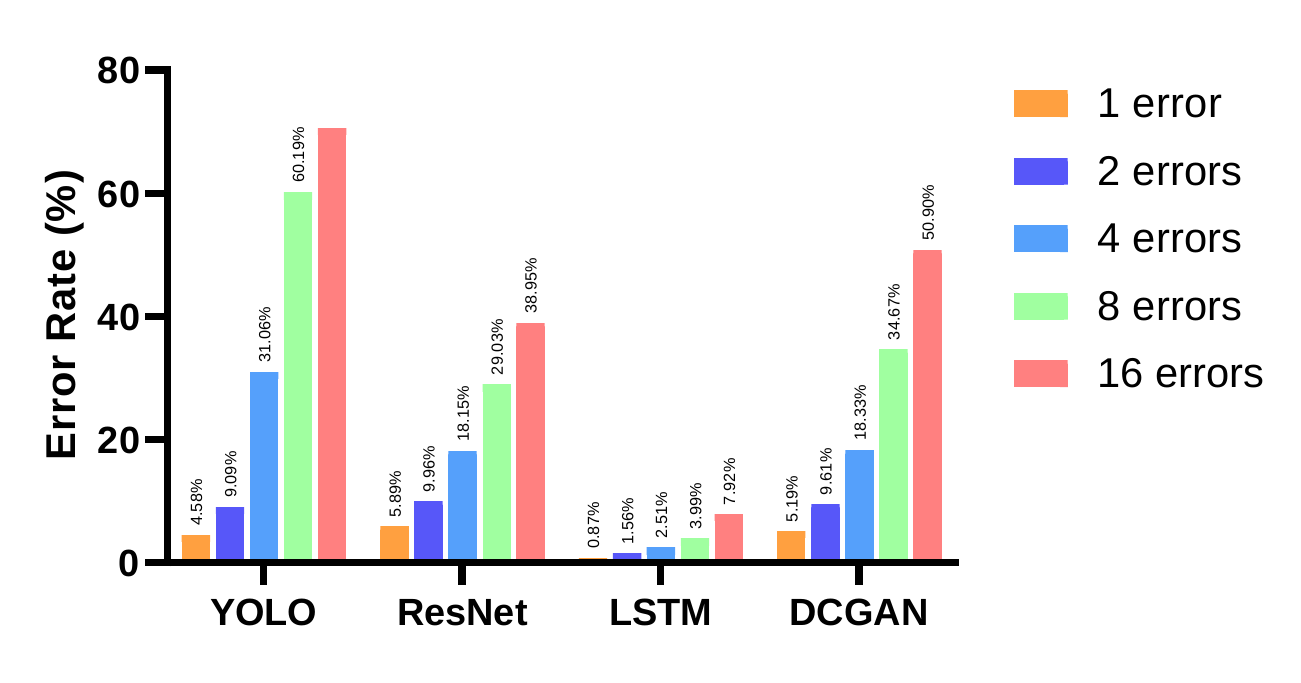}}
    \caption{Application error rate caused by single-bit random hardware error}
\label{fig:lab1-error-rate}
\vspace{-0.5em}
\end{figure}

First, we conducted single-bit random error injection 
experiments for different applications, and conducted 
20,000 runs for each application, to analyze the proportion 
of single-bit random error shielded in the system and the 
possible influence of single-bit random error on the 
accelerator system. Figure \ref{fig:lab1-error-rate} shows the 
percentage of application errors caused by a single-bit 
random hardware error. In the experiment, we found that 
more than 90\% of the errors were masked by software or 
hardware. However, while most errors are masked without 
impact to the operation, a single-bit error can lead to 
serious exceptions, including system halt, serious 
deviation in results, and so on. In addition, LSTM network 
has a better fault tolerance performance than other three 
networks, and the proportion of the influence caused by 
single error of the other three networks is about 5~7 times 
that of LSTM. According to our analysis, this is because 
LSTM network is smaller than other networks and uses less 
storage and computing resources.

\subsection{Effect of Error Number}
\begin{figure}
	\center{\includegraphics[width=0.85\linewidth]{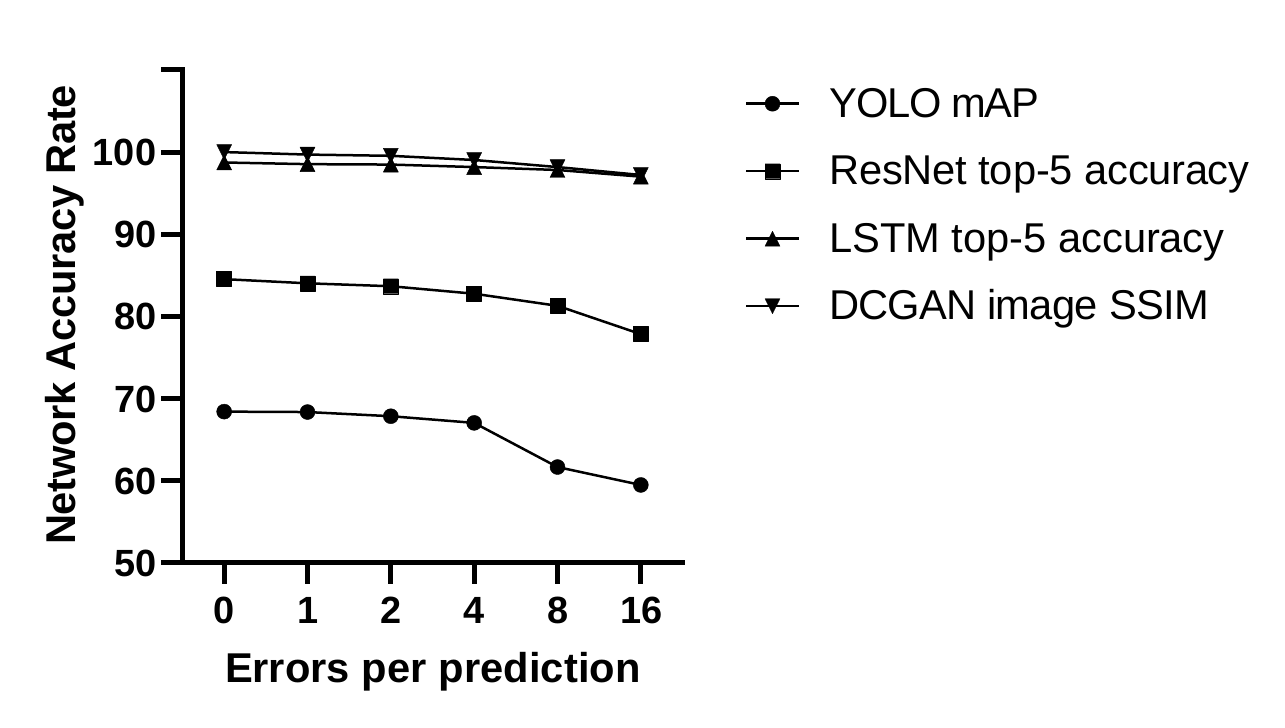}}
    \caption{Network accuracy versus error number}
\label{fig:lab2-network-accuracy}
\vspace{-0.5em}
\end{figure}

\begin{figure}
    \centering
    \subfigure[System halt]{\includegraphics[width=0.45\linewidth]{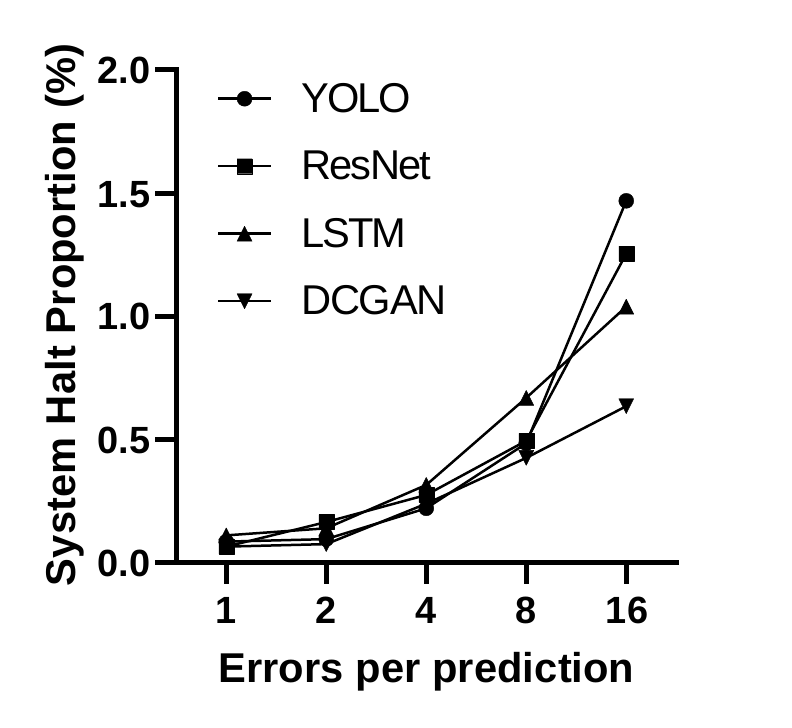}}
    \subfigure[Result with deviation]{\includegraphics[width=0.45\linewidth]{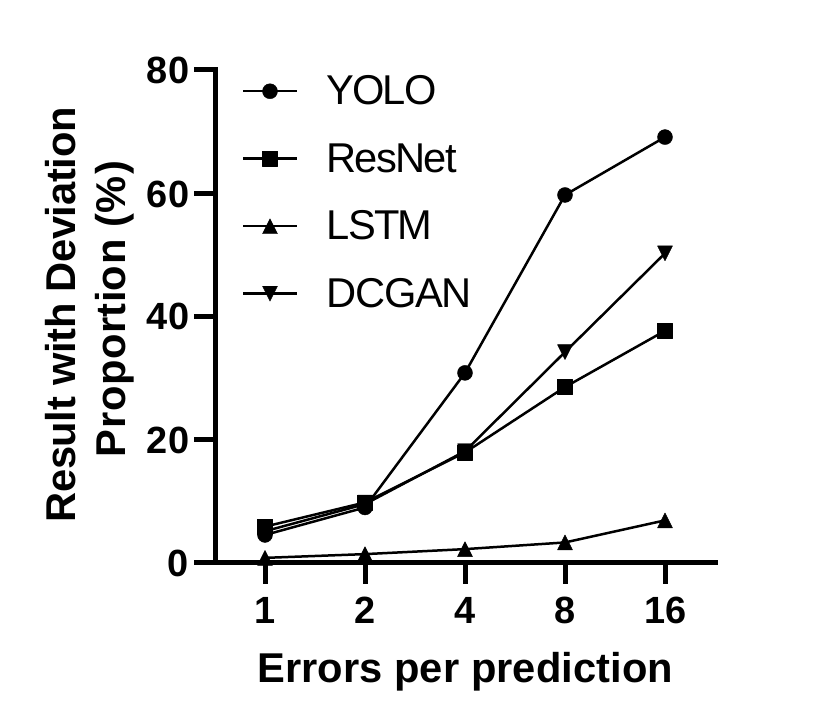}}
    \caption{Abnormal situation proportion versus error number}
\label{fig:lab2_2}
\vspace{-0.5em}
\end{figure}

We completed a single run with errors injected number of 
powers of 2, with 20,000 runs for each network. In general, 
the proportion of system halt and result with deviation 
increases with the number of errors, and the accuracy of 
application decreases. Figure \ref{fig:lab2_2} show the proportion of 
system halt and result deviation, and Figure \ref{fig:lab2-network-accuracy} shows the 
accuracy of each network. We found that the influence of 
multiple errors on the accelerator was obvious and far 
beyond our expectation. The Neural Network accelerator is 
still vulnerable to errors. By the time we injected 16 
errors in a single run, the system halt rate was more than 
1\%, means it took more time to restore the system than to 
run it. From the perspective of network accuracy, take YOLO 
system as an example, its mAP decreases by 8.95\% when it 
runs with 16 errors, means the application function of the 
system is also seriously affected.

The result with deviation proportion of different networks 
appears differently with the increase of error number. When 
16 errors are injected, LSTM network still has a small 
result deviation proportion due to its small network 
structure. The proportion of YOLO system is very high, with 
about 70\% of the results showing errors. ResNet is 
relatively low, with only about 35\% result deviation. 
DCGAN network is in between, and about 50\% of the results 
have numerical errors. We believe that the result deviation 
may be related to the structure of the output. The output 
of YOLO system contains more information such as object 
type, location and size of bounding box, etc., and the 
output of ResNet is simple sorting, while the simple output 
is obviously less susceptible to errors.

\subsection{Details of Result with Deviation}

\begin{figure*}
    \centering
    \subfigure[YOLO]{\includegraphics[height=0.12\textheight]{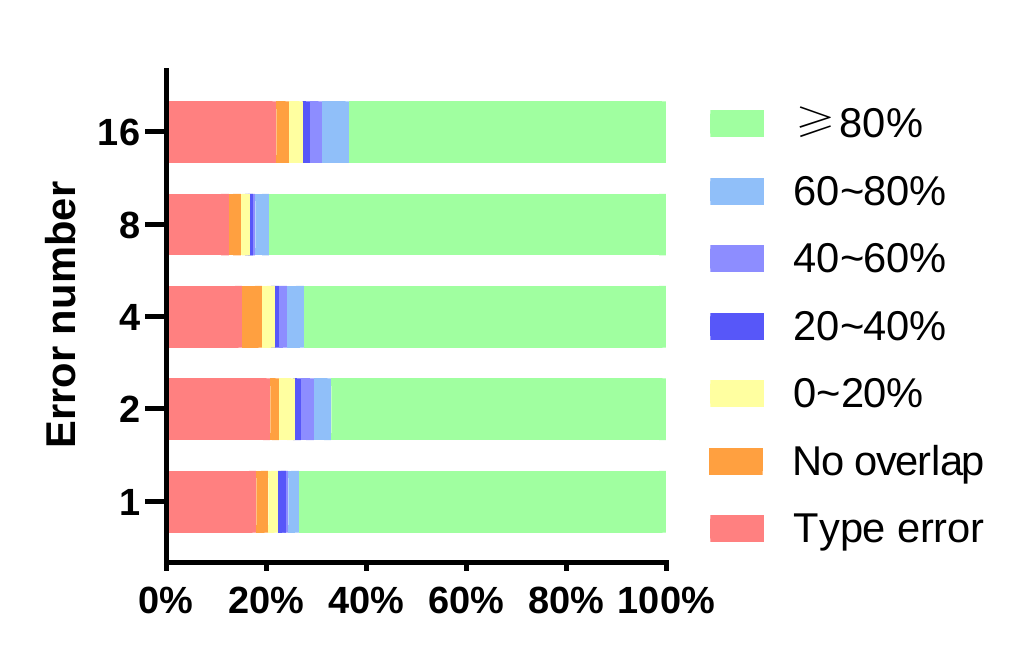}}
    \subfigure[ResNet]{\includegraphics[height=0.12\textheight]{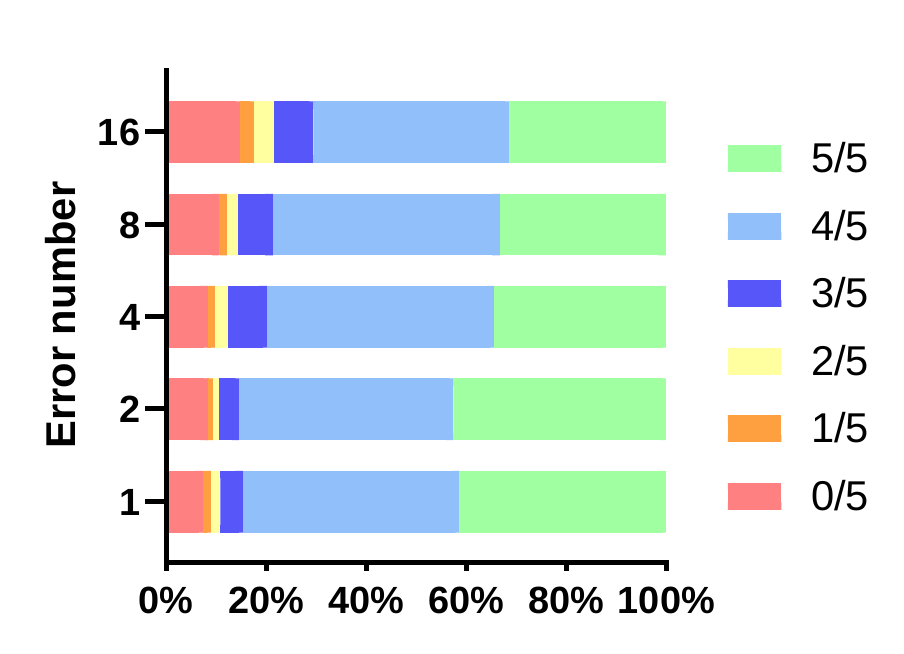}}
    \subfigure[LSTM]{\includegraphics[height=0.12\textheight]{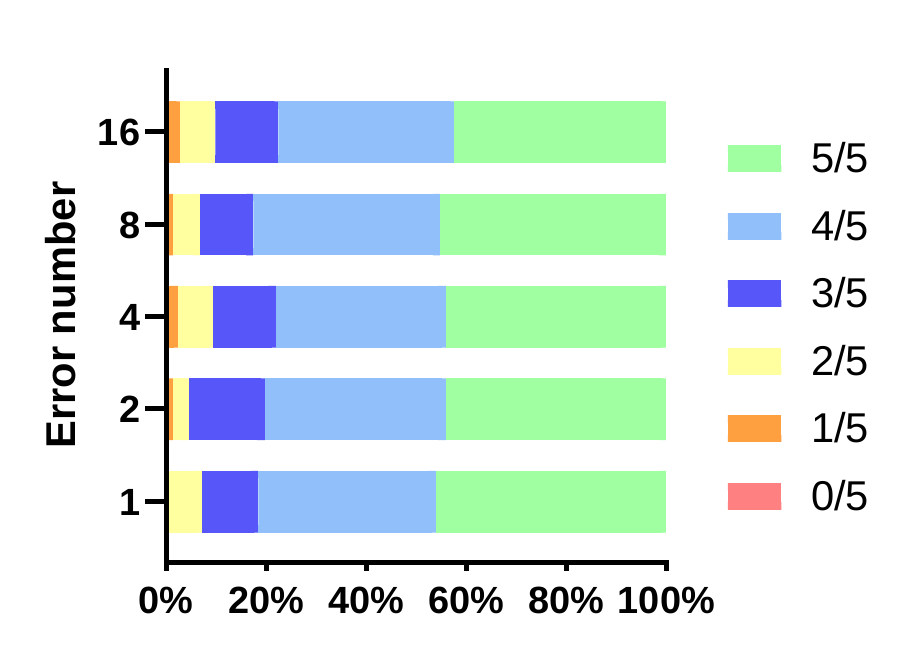}}
    \subfigure[DCGAN]{\includegraphics[height=0.12\textheight]{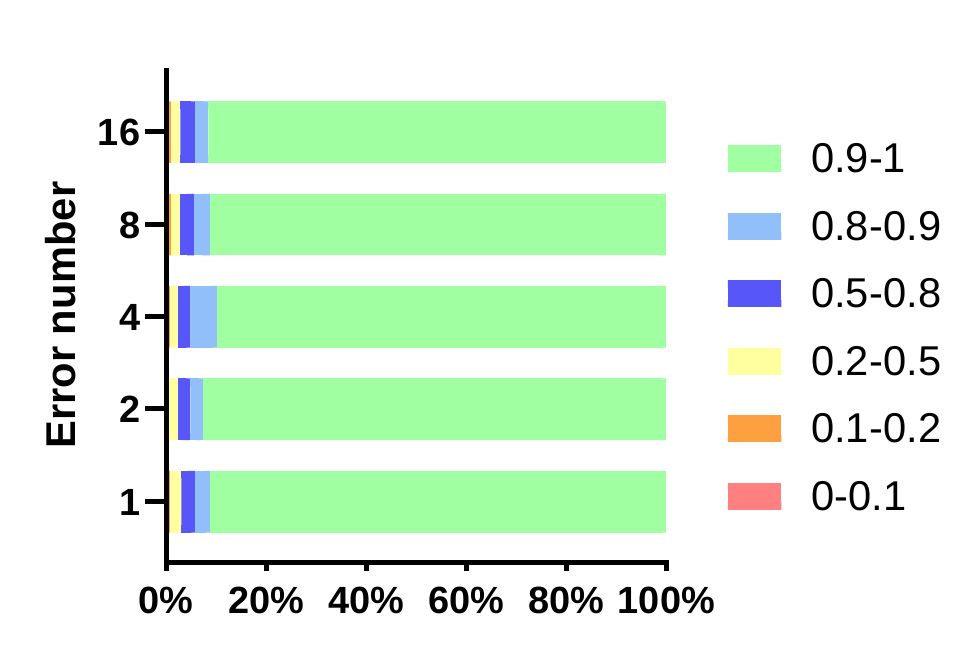}}
\caption{Distribution of result with deviation situations}
\label{fig:lab3}
\vspace{-0.5em}
\end{figure*}

\begin{figure*}
    \centering
    \subfigure[YOLO]{\includegraphics[height=0.10\textheight]{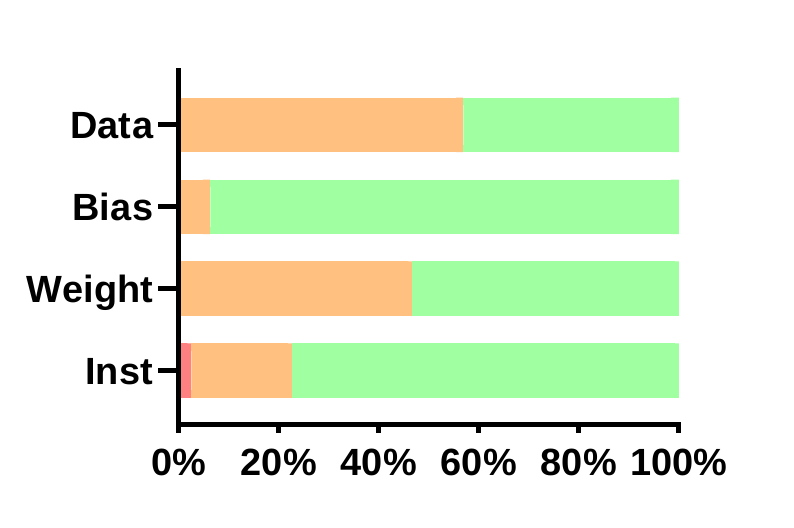}}
    \subfigure[ResNet]{\includegraphics[height=0.10\textheight]{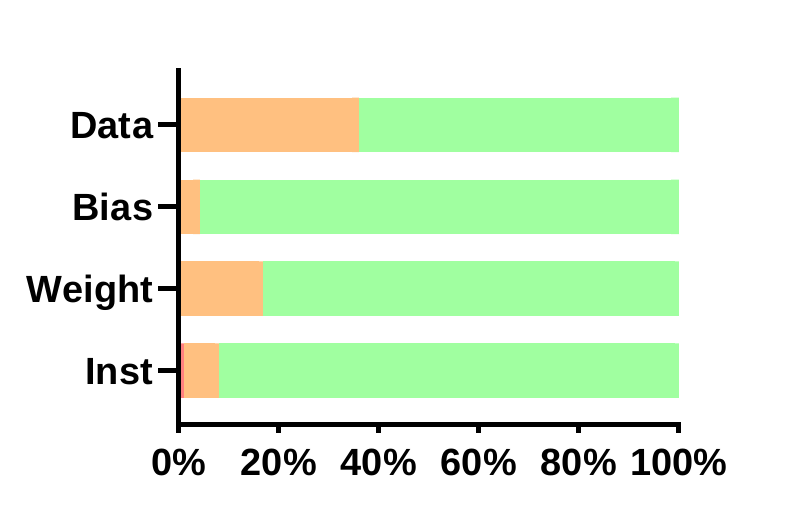}}
    \subfigure[LSTM]{\includegraphics[height=0.10\textheight]{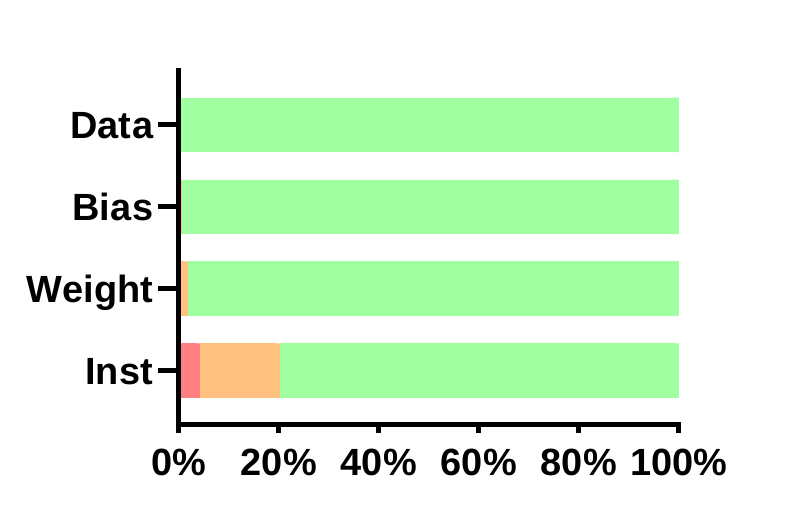}}
    \subfigure[DCGAN]{\includegraphics[height=0.10\textheight]{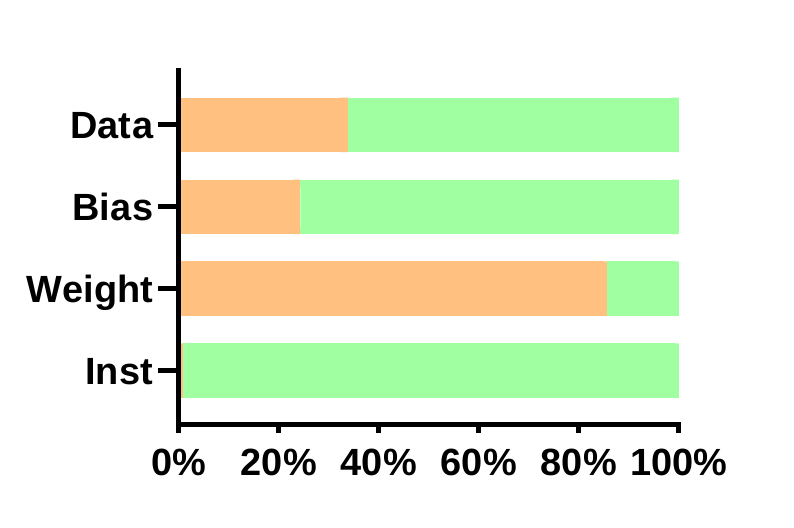}}
    \includegraphics[height=0.10\textheight]{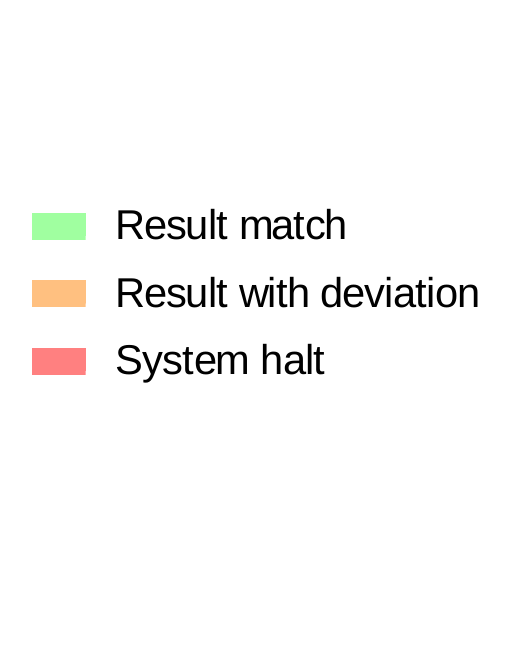}
\caption{Proportion of different error location}
\label{fig:lab4}
\vspace{-0.5em}
\end{figure*}

The results with errors are further classified and analyzed. In general, most of the errors are small, with a certain proportion of serious errors and relatively few moderate ones. We believe that most of the errors do not belong to the errors with global influence, but only affect one or several calculations. For example, a single value in the convolution kernel changes. Alternatively, a portion of the error is masked by subsequent calculations such as the max pooling layer. Some errors may have an impact on the control path or the reusable module, resulting in the accumulation of errors throughout the calculation; Or errors that cause serious deviations in the data, such as sign bit upset, can have serious consequences. Figure \ref{fig:lab3} shows the details of result distribution of result with deviation situations.

Specific to each network, about 70\% of the YOLO system's errors belong to the level of bounding box overlap ratio more than 80\%. The total of object type errors and non-overlapping boxes is about 20\%. The remaining 10\% or so is moderate errors. We think this is caused by the implementation of YOLO. YOLO divides the input images into a series of grid cells, and each grid cell is only responsible for one kind of object. Moreover, there is a binary judgment Pr(Object) whether there is a target or not in the confidence degree, which leads to more object type error cases. The bounding boxes that are given after object recognition based on relative wide and high are less sensitive.

More than 70\% of the errors in the ResNet and LSTM classification networks were small impact results of matching four or five items. We think this is because the output is SoftMax layer, which is less affected by the error, and the error is easy to be hidden when sorting, so the result only shows a small alter. However, with the increase of the number of errors, the proportion of serious errors of no matching item in ResNet results increased with the increase of the number of errors and exceeded 10\%. Serious errors cannot be hidden, and the networks fault tolerance for multiple errors is relatively limited.

The result of DCGAN is about 90\% of the results are small deviation level of more than 90\% of SSIM. Only a few of them have a large deviation. Combined with the image, in 90\% of the small deviation results compared with the standard output, basically no difference can be directly seen, and only a few large deviation results have serious distortion of the image. Considering that the output is picture information, these small deviations can be ignored without affecting the visual effect, we believe that DCGAN has relatively strong error tolerance.

\subsection{Effect of Error Location}
In this period, we conducted the experiment results of different error locations. We injected a single-bit random error into a designated location, and conducted multiple experiments to observe the performance of the network. By analyzing the influence of error in different locations on the accelerator, it can provide specific methods for the subsequent fault-tolerant design.

In general, the system is less affected by configuration memory errors, considering that the error injection of configuration memory is global, whose errors may not affect the system. Errors in configuration memory can cause system halt or result with deviation. Errors in BRAMs used for instruction buffer can cause system halt or result with deviation, while BRAMs used in other type buffers can only cause numerical deviations.

We focus on the analysis of the system halt situations caused by error in instruction buffer, which take about 20\% part of the system halt situations. Compared with the instruction before and after upset, the system halts caused by instruction errors includes three situations: instruction type changes, wrong instruction not defined, and the parameter in the operation is abnormal. Different instruction types of the original instruction are considered. Table \ref{tab:instruction change} and Table \ref{tab:original instruction type} shows the detail of instruction error. About 50\% of the system halt situations are caused by the error of DMA instructions. The abnormal access address or boundary violation caused by the abnormal parameters of DMA instruction will lead to the system halt. About 30\% are caused by AGU instruction errors, which result in abnormal in-chip control flow.

\begin{table}
    \centering
    \caption{Instruction change}
    \label{tab:instruction change}
    \begin{tabular}{cc}
        \toprule
            Situation & Percentage \\
        \midrule
            Error instruction undefined & 2.48\% \\
            Instruction type change & 7.45\% \\
            Abnormal parameter & 90.06\% \\
        \bottomrule
    \end{tabular}
\vspace{-1em}
\end{table}

\begin{table}
    \centering
    \caption{Original instruction type}
    \label{tab:original instruction type}
    \begin{tabular}{cc}
        \toprule
            type & Percentage \\
        \midrule
            DMA & 55.90\% \\
            AGU & 32.92\% \\
            others & 11.18\% \\
        \bottomrule
    \end{tabular}
\vspace{-1em}
\end{table}

The proportions of errors in instruction buffer led to system halt in ResNet, YOLO and LSTM, reached 1.15\%, 2.55\% and 4.25\% respectively, seriously affecting the proper application of the system. For result with deviation case, in the experiment of YOLO, about half of the result deviation situations caused by instruction buffer errors were serious object type errors. In the experiment of ResNet, a large number of mismatches were also caused. DCGAN network is limited by instruction errors, because the instruction sequence length of this network is very short, only 2\% of the instruction buffer is used, while the instruction sequence of other networks uses instruction buffer of 50\%\~{}80\%. Combined with the result with deviation and system halt case, we propose that instruction buffer needs to be strengthened in the fault-tolerant design.

Errors in the buffers used in data-flow, such as weights, data, and bias buffer, do not cause system halt, only may cause result deviations. In general, data and weights are more sensitive than bias. Taking YOLO system as an example, the proportion of result deviation caused by single error in weight, data and bias buffer is 48.75\%, 56.95\% and 6.35\% respectively. Horizontal comparison shows that each network has different sensitivity to different buffer errors, as shown in figure {}.

\subsection{Input-Related Error}
In the above experiments, we used different input data for 
experiments, and we verified the relationship between 
errors and input data in this set of experiments. We test 
different input data using the same error. Whether a 
hardware error causes an application error exists in two 
ways, depending on the input data or not. Errors unrelated 
to input data, fixed to cause system halt or result 
deviation, or be masked, that is, different input data will 
not influence the classification of the result. The other 
part of the errors is input data related, which can be 
shown as replacing different input data, result match 
situation and result with deviation situation both appear. 
In other words, different input data has different 
sensitivity to an input-related error. We believe that this 
is caused by structures which error affected. 
Input-unrelated errors may affect the control-flow of the 
system, the bus, etc. These structures are used in every 
operation, and the errors will not be masked by subsequent 
calculations. Input-related errors may affect the relevant 
data in the data-flow, and some errors may be masked in 
the subsequent calculation. Due to the large number of 
experiments, we only observed this phenomenon without 
further study on its proportion and distribution.

\section{Related work} \label{sec:relatedwork}
Deep learning accelerators that ensure both high
performance and energy efficiency of neural network are 
increasingly adopted in various computing devices including 
IoT devices, mobile phones and cloud. While the accelerators 
fabricated with the latest semiconductor technology is 
susceptible to the manufacturing defects and abnormal 
processes due to the small transistor feature size, the 
number of hardware faults grows accordingly. The faults may 
further lead to computing errors, dramatic prediction accuracy 
loss and even system stall when the faults are not handled 
appropriately. For some of the mission-critical applications, 
the prediction mistakes may even cause catastrophic failures.

To gain insight of fault tolerance of neural network accelerators, 
the authors in \cite{Li2017} investigated the influence of data types, 
values, data reuses, and types of layers on the resilience of DNN 
accelerators through experiments on a DNN simulator. error resilience of xxx with simulation. Different from xxx,
the works in xxx mainly focus on the difference of fault tolerance using 
different data types. Some of the researchers try 
The reliability of the neural network accelerators 
While neural networks with large amount of redundant computing 
is considered to resilient, Reliability of neural network accelerators become critical 
to resilient neural network execution. 
1) Reliability espeically FPGA based reliability problem
2) The prevalence of deep learning neural networks provokes 
the development of convolutional neural network (CNN) accelerators 
for both higher performance and energy efficiency. 
existing fault-tolerant CNN accelerator works,

Data type analysis with simulation, error propagation analysis
Retraining to improve fault tolerance and tolerate the computing errors
(Change the neural network model) Computing array based fault model
Relax the design constraints and have the accelerator 
obtain advantageous design trade-off between precision and performance.DAC'18 work
Basically, the analysis focuses on the computing of the neural network.
Hardware structure is not discussed in detail. 
Simulation is the major approach.
lack of system analysis on a CPU-CNN accelerator. 
FPGA based analysis is not covered.
Prior FPGA based reliability such as soft processors etc. It is 
not quite relevant.
Neural network inherent fault tolerance.
Error analysis on a running system remains not explored. 

FPGA is a widely used hardware platform, and there are many 
designs and methods for error injection on FPGAs 
\cite{Ebrahimi2014A}\cite{Lopez2007A}
\cite{Harward2015Estimating}\cite{Tarrillo2015Multiple}. 
FPGA-based (also known as emulation-based) error analysis 
is widely used in soft processor sensitivity analysis and 
other scenarios. The FPGA-based fault injection techniques 
have good controllability, observability and ideal speed.

FPGA-based fault injection techniques can be divided into 
two categories: reconfiguration-based techniques and 
instrument-based techniques. The reconfiguration-based 
techniques rely on the internal mechanism of FPGAs. These 
techniques use complete or partial reconfiguration to 
change the configuration bit of the FPGA device to apply 
the target fault model to the desired fault location. The 
reconfiguration process is a speed bottleneck for 
reconfiguration-based techniques. In instrument-based 
technology, fault injection modules called saboteur are 
added to each fault point, and fault is injected by 
activating the saboteurs. Instrument-based techniques have 
higher speed than reconfiguration-based techniques, while 
implementing the saboteurs increase the area of the circuit. 
We shall design the saboteurs ourselves for 
instrument-based techniques, and insert them into the 
design under test.
\section{Conclusion} \label{sec:conclusion}

With the wide application of neural network in many 
scenarios with high safety requirements like autonomous 
driving, the fault-tolerant performance of neural network 
accelerators has received more attention. In this paper, we 
designed a fault emulation and analysis system for neural 
network accelerator based on SRAM-based FPGA, and completed 
a series of experiments on this basis. We find that the 
fault tolerance of neural network accelerator is poor, and 
analyze the misrepresentation of different numbers, 
different network applications and different error 
locations. These analyses provide a reference for the 
following fault-tolerant design.

\bibliographystyle{IEEEtran}
\bibliography{refs}

\begin{thebibliography}{10}
\providecommand{\url}[1]{#1}
\csname url@samestyle\endcsname
\providecommand{\newblock}{\relax}
\providecommand{\bibinfo}[2]{#2}
\providecommand{\BIBentrySTDinterwordspacing}{\spaceskip=0pt\relax}
\providecommand{\BIBentryALTinterwordstretchfactor}{4}
\providecommand{\BIBentryALTinterwordspacing}{\spaceskip=\fontdimen2\font plus
\BIBentryALTinterwordstretchfactor\fontdimen3\font minus
  \fontdimen4\font\relax}
\providecommand{\BIBforeignlanguage}[2]{{%
\expandafter\ifx\csname l@#1\endcsname\relax
\typeout{** WARNING: IEEEtran.bst: No hyphenation pattern has been}%
\typeout{** loaded for the language `#1'. Using the pattern for}%
\typeout{** the default language instead.}%
\else
\language=\csname l@#1\endcsname
\fi
#2}}
\providecommand{\BIBdecl}{\relax}
\BIBdecl

\bibitem{Gatys_2016_CVPR}
L.~A. Gatys, A.~S. Ecker, and M.~Bethge, ``Image style transfer using
  convolutional neural networks,'' in \emph{The IEEE Conference on Computer
  Vision and Pattern Recognition (CVPR)}, June 2016.

\bibitem{Collobert:2008:UAN:1390156.1390177}
\BIBentryALTinterwordspacing
R.~Collobert and J.~Weston, ``A unified architecture for natural language
  processing: Deep neural networks with multitask learning,'' in
  \emph{Proceedings of the 25th International Conference on Machine Learning},
  ser. ICML '08.\hskip 1em plus 0.5em minus 0.4em\relax New York, NY, USA: ACM,
  2008, pp. 160--167. [Online]. Available:
  \url{http://doi.acm.org/10.1145/1390156.1390177}
\BIBentrySTDinterwordspacing

\bibitem{Chen_2015_ICCV}
C.~Chen, A.~Seff, A.~Kornhauser, and J.~Xiao, ``Deepdriving: Learning
  affordance for direct perception in autonomous driving,'' in \emph{The IEEE
  International Conference on Computer Vision (ICCV)}, December 2015.

\bibitem{chen2014diannao}
T.~Chen, Z.~Du, N.~Sun, J.~Wang, C.~Wu, Y.~Chen, and O.~Temam, ``Diannao: A
  small-footprint high-throughput accelerator for ubiquitous
  machine-learning,'' in \emph{ACM Sigplan Notices}, vol.~49, no.~4.\hskip 1em
  plus 0.5em minus 0.4em\relax ACM, 2014, pp. 269--284.

\bibitem{chen2014dadiannao}
Y.~Chen, T.~Luo, S.~Liu, S.~Zhang, L.~He, J.~Wang, L.~Li, T.~Chen, Z.~Xu,
  N.~Sun \emph{et~al.}, ``Dadiannao: A machine-learning supercomputer,'' in
  \emph{Proceedings of the 47th Annual IEEE/ACM International Symposium on
  Microarchitecture}.\hskip 1em plus 0.5em minus 0.4em\relax IEEE Computer
  Society, 2014, pp. 609--622.

\bibitem{chen2016eyeriss}
Y.-H. Chen, J.~Emer, and V.~Sze, ``Eyeriss: A spatial architecture for
  energy-efficient dataflow for convolutional neural networks,'' in \emph{ACM
  SIGARCH Computer Architecture News}, vol.~44, no.~3.\hskip 1em plus 0.5em
  minus 0.4em\relax IEEE Press, 2016, pp. 367--379.

\bibitem{Zhang:2015:OFA:2684746.2689060}
\BIBentryALTinterwordspacing
C.~Zhang, P.~Li, G.~Sun, Y.~Guan, B.~Xiao, and J.~Cong, ``Optimizing fpga-based
  accelerator design for deep convolutional neural networks,'' in
  \emph{Proceedings of the 2015 ACM/SIGDA International Symposium on
  Field-Programmable Gate Arrays}, ser. FPGA '15.\hskip 1em plus 0.5em minus
  0.4em\relax New York, NY, USA: ACM, 2015, pp. 161--170. [Online]. Available:
  \url{http://doi.acm.org/10.1145/2684746.2689060}
\BIBentrySTDinterwordspacing

\bibitem{7551397}
S.~{Han}, X.~{Liu}, H.~{Mao}, J.~{Pu}, A.~{Pedram}, M.~A. {Horowitz}, and W.~J.
  {Dally}, ``Eie: Efficient inference engine on compressed deep neural
  network,'' in \emph{2016 ACM/IEEE 43rd Annual International Symposium on
  Computer Architecture (ISCA)}, June 2016, pp. 243--254.

\bibitem{238315}
P.~W. {Protzel}, D.~L. {Palumbo}, and M.~K. {Arras}, ``Performance and
  fault-tolerance of neural networks for optimization,'' \emph{IEEE
  Transactions on Neural Networks}, vol.~4, no.~4, pp. 600--614, July 1993.

\bibitem{Reagen2018}
B.~Reagen, U.~Gupta, L.~Pentecost, P.~Whatmough, S.~K. Lee, N.~Mulholland,
  D.~Brooks, and G.-Y. Wei, ``{Ares: A framework for quantifying the resilience
  of deep neural networks},'' pp. 1--6, 2018.

\bibitem{Kausar2016Artificial}
F.~{Kausar} and P.~{Aishwarya}, ``Artificial neural network: Framework for
  fault tolerance and future,'' in \emph{2016 International Conference on
  Electrical, Electronics, and Optimization Techniques (ICEEOT)}, March 2016,
  pp. 648--651.

\bibitem{Li2018TensorFI}
G.~Li, K.~Pattabiraman, and N.~Debardeleben, ``Tensorfi: A configurable fault
  injector for tensorflow applications,'' in \emph{2018 IEEE International
  Symposium on Software Reliability Engineering Workshops (ISSREW)}, 2018.

\bibitem{Li2017}
G.~Li, S.~K.~S. Hari, M.~Sullivan, T.~Tsai, K.~Pattabiraman, J.~Emer, and S.~W.
  Keckler, ``{Understanding error propagation in deep learning neural network
  (DNN) accelerators and applications},'' pp. 1--12, 2017.

\bibitem{Salami2018}
\BIBentryALTinterwordspacing
B.~Salami, O.~Unsal, and A.~Cristal, ``{On the Resilience of RTL NN
  Accelerators: Fault Characterization and Mitigation},'' 2018. [Online].
  Available: \url{http://arxiv.org/abs/1806.09679}
\BIBentrySTDinterwordspacing

\bibitem{li2021novelty}
S.~Li, J.~Niu, and Z.~Li, ``Novelty detection of cable-stayed bridges based on
  cable force correlation exploration using spatiotemporal graph convolutional
  networks,'' \emph{Structural Health Monitoring}, p. 1475921720988666, 2021.

\bibitem{shi2021fault}
C.~Shi, Y.~Ren, H.~Tang, and L.~R. Mupfukirei, ``A fault diagnosis method for
  an electro-hydraulic directional valve based on intrinsic mode functions and
  weighted densely connected convolutional networks,'' \emph{Measurement
  Science and Technology}, vol.~32, no.~8, p. 084015, 2021.

\bibitem{UG953}
{Xilinx Inc.}, ``Vivado design suite 7 series fpga and zynq-7000 soc libraries
  guide,''
  \url{http://www.xilinx.com/support/documentation/sw_manuals/xilinx2017_4/ug953-vivado-7series-libraries.pdf},
  Dec 2018, \ UG953(v2017.4).

\bibitem{PG134}
------, ``Axi hwicap v3.0 logicore ip product guide,''
  \url{http://www.xilinx.com/support/documentation/ip_documentation/axi_hwicap/v3_0/pg134-axi-hwicap.pdf},
  Oct 2016, \ PG134.

\bibitem{UG470}
------, ``7 series fpgas configuration user guide,''
  \url{http://www.xilinx.com/support/documentation/user_guides/ug470_7Series_Config.pdf},
  Aug 2018, \ UG470 (v1.13.1).

\bibitem{redmon2016yolo9000}
J.~Redmon and A.~Farhadi, ``Yolo9000: Better, faster, stronger,'' \emph{arXiv
  preprint arXiv:1612.08242}, 2016.

\bibitem{He_2016_CVPR}
K.~He, X.~Zhang, S.~Ren, and J.~Sun, ``Deep residual learning for image
  recognition,'' in \emph{The IEEE Conference on Computer Vision and Pattern
  Recognition (CVPR)}, June 2016.

\bibitem{sak2014long}
H.~Sak, A.~Senior, and F.~Beaufays, ``Long short-term memory recurrent neural
  network architectures for large scale acoustic modeling,'' in \emph{Fifteenth
  annual conference of the international speech communication association},
  2014.

\bibitem{radford2015unsupervised}
A.~Radford, L.~Metz, and S.~Chintala, ``Unsupervised representation learning
  with deep convolutional generative adversarial networks,'' \emph{arXiv
  preprint arXiv:1511.06434}, 2015.

\bibitem{Ebrahimi2014A}
M.~Ebrahimi, A.~Mohammadi, A.~Ejlali, and S.~G. Miremadi, ``A fast, flexible,
  and easy-to-develop fpga-based fault injection technique,''
  \emph{Microelectronics Reliability}, vol.~54, no.~5, pp. 1000--1008, 2014.

\bibitem{Lopez2007A}
C.~Lopez-Ongil, L.~Entrena, M.~Garcia-Valderas, M.~Portela, and F.~Munoz, ``A
  unified environment for fault injection at any design level based on
  emulation,'' \emph{IEEE Transactions on Nuclear Science}, vol.~54, no.~4, pp.
  946--950, 2007.

\bibitem{Harward2015Estimating}
N.~A. Harward, M.~R. Gardiner, L.~W. Hsiao, and M.~J. Wirthlin, ``Estimating
  soft processor soft error sensitivity through fault injection,'' in
  \emph{IEEE International Symposium on Field-programmable Custom Computing
  Machines}, 2015.

\bibitem{Tarrillo2015Multiple}
J.~Tarrillo, J.~Tonfat, L.~Tambara, F.~L. Kastensmidt, and R.~Reis, ``Multiple
  fault injection platform for sram-based fpga based on ground-level radiation
  experiments,'' in \emph{Test Symposium}, 2015.

\end{thebibliography}

\end{document}